\documentclass{article}
\usepackage{spconf,amsmath,epsfig,cite,multirow}

\let\OLDthebibliography\thebibliography
\renewcommand\thebibliography[1]{
  \OLDthebibliography{#1}
  \setlength{\parskip}{0pt}
  \setlength{\itemsep}{0pt plus 0.3ex}
}

\begin{document}\sloppy

\def\x{{\mathbf x}}
\def\L{{\cal L}}

\title{Learning content and context with language bias for Visual Question Answering}
%
\name{Chao Yang$^{\ast}$, Su Feng$^{\ast}$, Dongsheng Li$^{\dagger}$, Huawei Shen$^{\ddagger}$, Guoqing Wang$^{\ast}$, Bin Jiang$^{\ast}$}
    
\address{$^{\ast}$College of Computer Science and Electronic Engineering, Hunan University\\ 
          \{ yangchaoedu, hnufs, wgqbeam, jiangbin \}@hnu.edu.cn\\
          $^{\dagger}$Microsoft Research Asia, dongsli@microsoft.com\\
          $^{\ddagger}$Insitute of Computing Technology, Chinese Academy of Science, shenhuawei@ict.ac.cn}


\maketitle

\begin{abstract}
Visual Question Answering (VQA) is a challenging multimodal task to answer questions about an image. Many works concentrate on how to reduce language bias which makes models answer questions ignoring visual content and language context. However, reducing language bias also weakens the ability of VQA models to learn context prior. To address this issue, we propose a novel learning strategy named CCB, which forces VQA models to answer questions relying on Content and Context with language Bias. Specifically, CCB establishes Content and Context branches on top of a base VQA model and forces them to focus on local key content and global effective context respectively. Moreover, a joint loss function is proposed to reduce the importance of biased samples and retain their beneficial influence on answering questions. Experiments show that CCB outperforms the state-of-the-art methods in terms of accuracy on VQA-CP v2.
\end{abstract}
\section{Introduction}
\label{sec:intro}
Visual Question Answering (VQA)~\cite{antol2015vqa} is an attractive task spanning computer vision and natural language processing. Given an image and a textual question, the task aims to generate an answer in natural language. Ideally, a VQA model requires a deep understanding of semantic information from visual and textual modalities. However, many recent works~\cite{goyal2017making,agrawal2018don} have pointed out that most existing VQA models answer questions strongly relying on language bias, which is the superficial statistical correlation between question and answer. 

%
\begin{figure}[htb]
\begin{minipage}[b]{1.0\linewidth}
  \centering
  \centerline{\epsfig{figure=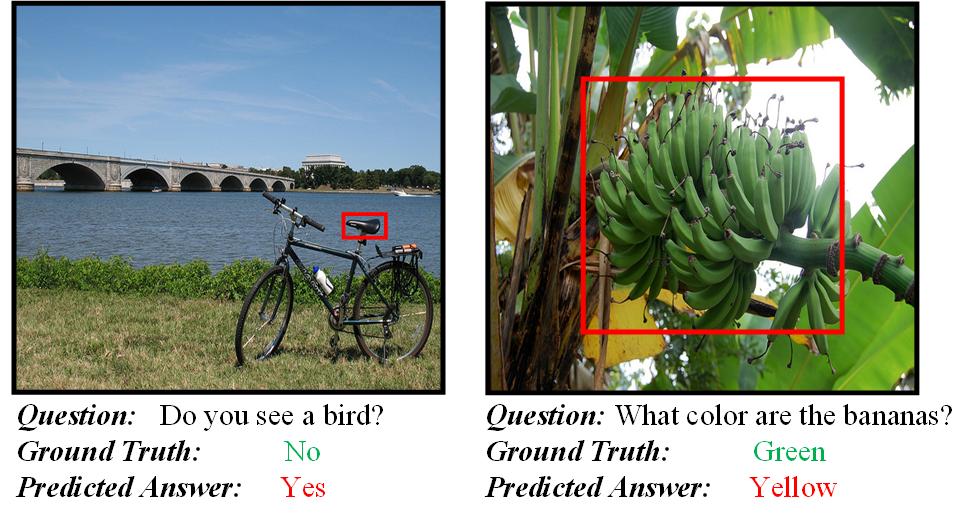,width=8.5cm}}
\end{minipage}
\caption{Example of the VQA language bias. When asked ``Do you see a ...?'', the model tends to simply answer ``yes'' based on the salient object in the picture. For questions like ``What color ...?'', even if the model pays attention to the region of ``the bananas'', it will give the most common answer ``yellow''.}
\label{fig:res}
\end{figure}

Specifically, the VQA language bias~\cite{niu2020counterfactual} is generally divided into the ``visual priming bias''~\cite{antol2015vqa,goyal2017making} and the ``language prior''~\cite{agrawal2018don,ramakrishnan2018overcoming}. As shown in Figure 1, models usually tend to focus on visual salient content and ignore language context. For questions like ``Do you see a...'', as long as  models simply answer ``yes'', they can achieve a high accuracy in VQA v1 datasets~\cite{antol2015vqa}. Even if models have a correct visual grounding, they may directly generate the most common answers ({\em e.g.}, ``yellow'' bananas are more common than ``green'' ones in the training set). This shortcut severely limits the application of VQA in the real world, where the statistical distribution of questions and answers may be significantly different from the VQA dataset. In addition, the tendency to capture language bias seriously affects the estimation of model capability.

To alleviate language bias, VQA-CP~\cite{agrawal2018don} dataset is established with different question-answer distributions between the training and test splits. Many VQA models~\cite{yang2016stacked,anderson2018bottom} are found relying overly on language bias to answer questions and  have a significant decrease in accuracy on VQA-CP dataset. Existing approaches can be categorized as extra-based methods and non-extra-based methods. Extra-based methods are often intuitive: they introduce extra human supervision~\cite{selvaraju2019taking, wu2019self} and auxiliary tasks (visual grounding, image caption, {\em etc.})~\cite{selvaraju2019taking, wu2019self, qiao2020rankvqa, zhu2020overcoming} to increase the image dependency. In non-extra-based methods, the prevailing solutions to language bias are ensemble-based methods~\cite{ramakrishnan2018overcoming,cadene2019rubi,clark2019don}, which adopt an auxiliary QA branch and specific learning strategies. However, the above-mentioned methods all explore how to weaken the influence of language bias on the model, ignoring that language bias can also help models learn context prior ({\em e.g.}, ``what color'')~\cite{niu2020counterfactual,chen2020counterfactual}. 

To effectively utilize language bias in VQA task, we propose a novel CCB learning strategy to learn Content and Context with language Bias. Firstly, we build our Content and Context branches on top of Up-Down~\cite{anderson2018bottom} model, which consists of three components: extracting multimodal features, fusing these features to generate a joint vector and predicting the answer. The Content and Context branches share the same inputs but have different fusion methods. The Content branch concentrates on the local key information and the Context branch focuses on the global effective information in the image and question. Secondly, we design two different optimization objectives with language bias, to reduce the ``bad'' statistical prior in the Content branch and keep the ``good'' context prior in the Context branch. The two objectives can make the predictions of two branches be more sensitive to visual content and language context separately. Finally, the model produces an answer distribution under the joint influence of the two branches, avoiding that the model only rely on one-sided information to answer questions. Moreover, by constructing a joint loss function to optimize the Content branch, the Context branch and the joint prediction simultaneously, we achieve a trade-off between eliminating language bias and acquiring prior knowledge to better facilitate VQA task.

To summarize, our contricutions are as follows:

\begin{itemize}
\item We propose CCB, a novel learning strategy to disentangle language bias by building the Content and the Context branches. To our best knowledge, this is the first attempt to utilize language bias correctly instead of simply reducing, which makes the model more dependent on visual content and language context to answer questions.
\item We design a joint loss function to optimize the Content branch, the Context branch and the final answer prediction. It effectively helps the model overcome the statistical prior and retain the context prior simultaneously.
\item The experimental results show that our approach increases the overall accuracy from 52.05\% to 57.99\% on VQA-CP v2. Particularly, CCB achieves the state-of-the-art performance of 59.12\% with the CSS~\cite{chen2020counterfactual} training scheme.
\end{itemize}

\section{Related work}

In recent years, VQA task has attracted widespread attention~\cite{antol2015vqa,yang2016stacked,anderson2018bottom}, with the increasing demand for multimodal information understanding. Meanwhile, many studies~\cite{goyal2017making,agrawal2018don,niu2020counterfactual} have found that language bias in VQA, the strong statistical correlation between question and answer severely limits the generalization and understanding capabilities of VQA models for multimodal information. In response to the situation, VQA v2 dataset~\cite{goyal2017making} and VQA-CP dataset~\cite{agrawal2018don} are proposed. Most existing models~\cite{yang2016stacked,anderson2018bottom} have a significant decline in accuracy on the VQA-CP dataset, which has different distributions of QA pairs between the training set and the test set. Many methods have been proposed to solve this problem, and they can be roughly divided into two categories: extra-based methods and non-extra-based methods.

\subsection{Extra-based methods}

In extra-based methods, depending on extra human supervision is the most straightforward solution to increase the image dependency. HINT~\cite{selvaraju2019taking} optimizes the alignment between human attention maps~\cite{das2017human} and gradient-based network importance to reduce reliance on language bias. By human textual explanations~\cite{huk2018multimodal}, SCR~\cite{wu2019self} criticizes the sensitivity of incorrect answers to the influential object  for the same purpose. In order to better guide models to get answers, some extra auxiliary tasks are added to the existing VQA models. To some extends, SCR and HINT also are seen as adding extra visual grounding task to guide VQA models. RankVQA~\cite{qiao2020rankvqa} adds a image caption generator for reranking candidate answers from typical VQA model~\cite{anderson2018bottom}. SSL-VQA~\cite{zhu2020overcoming} introduces Question-Image correlation estimation as auxiliary task to overcome language prior in a self-supervised learning framework. In addition, as causal inference has gradually attracted attention~\cite{niu2020counterfactual,tang2020unbiased}, ~\cite{chen2020counterfactual} proposes a Counterfactual Samples Synthesizing training scheme to generate extra samples. Typically, extra-based methods can achieve better performance than non-extra methods, since they allow models to receive extra supervision, understand the task from an extra angle, or ``see'' extra samples.

\begin{figure*}[htb]
\begin{minipage}[b]{1.0\linewidth}
  \centering
 \centerline{\epsfig{figure=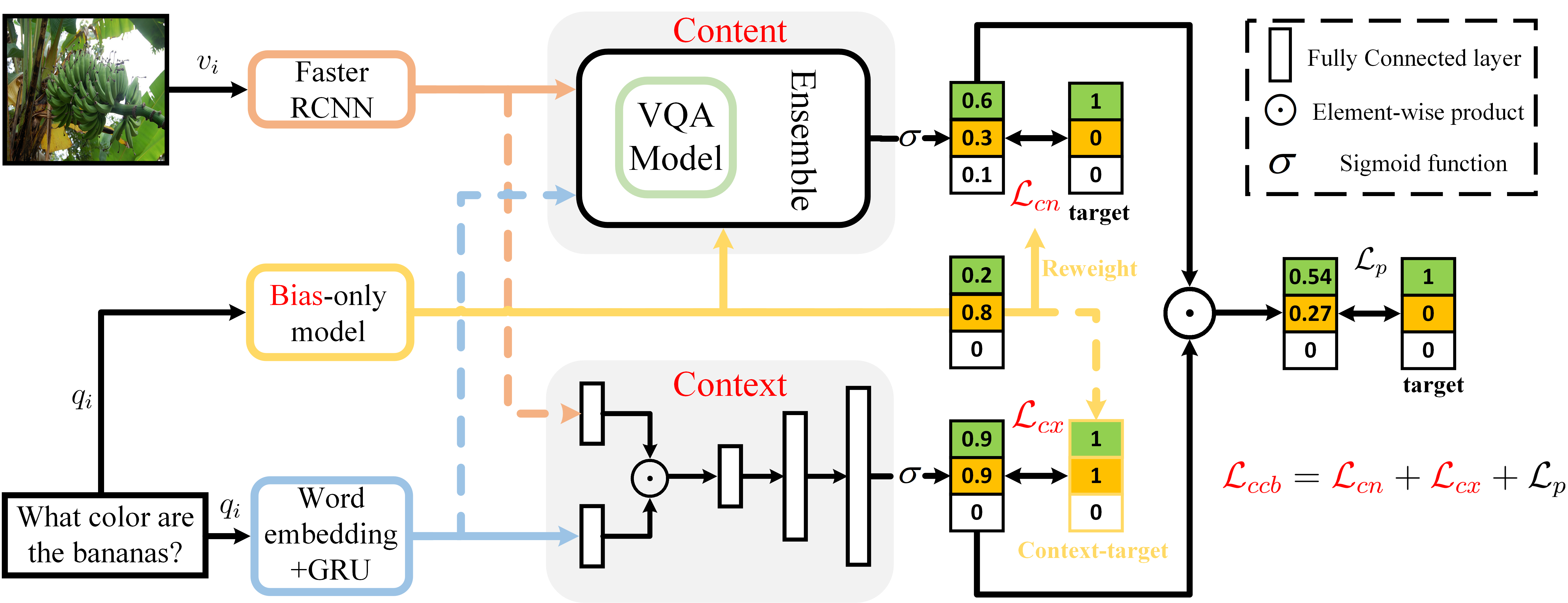,width=12cm}}
\end{minipage}
\caption{The overview of CCB learning strategy. It contains the Content branch, the Context branch and a joint loss function $\mathcal{L}_{ccb}$. Firstly, the two branches predict answers respectively through the local crucial content and global non-crucial context captured from multimodal features. Secondly, The final prediction is obtained under the joint influence of the two branches. Finally, we construct a joint loss function with language bias to jointly optimize the two branches and the final prediction.}
\label{fig:res}
\end{figure*}

\subsection{Non-extra-based methods}

Different from extra-based methods, non-extra-based methods does not introduce information outside VQA datasets. Currently, the mainstream methods to alleviate language bias are ensemble-based methods: they introduce a branch model and conduct joint training with the target VQA model by specific learning strategies. In adversarial training scheme, AReg~\cite{ramakrishnan2018overcoming} trains the VQA model and the question-only model which captures language bias by sharing the same question encoder. But this training scheme also affects the representations of questions and the stability of training~\cite{grand2019adversarial}. Rubi~\cite{cadene2019rubi} block the back propagation from the question-only model to the question encoder, and fuse the predicted answer distributions of the two branches to construct the loss function. Instead of the question-only model, LMH~\cite{clark2019don} introduces a pre-trained bias-only model to enhance the robustness of VQA model in a more targeted manner. In general, the current ensemble-based methods try to add an auxiliary branch to capture language bias so as to weaken its effect. However, this also impairs part of models’ ability in learning context prior. For achieving a balance, we propose CCB learning strategy.

\section{Methods}

\subsection{The Paradigm of VQA}

The common formulation of VQA task is often considered as a multi-class classification problem. Specifically, given a VQA dataset $D$ consisting of N triplets $(v_i,q_i,a_i)_{i\in[1,N]}$, where $v_i\in\mathcal{V}$, $q_i\in\mathcal{Q}$ and $a_i\in\mathcal{A}$ represent the image, the question and the answer for the $i_{th}$ instance, the VQA model aims to implement a function $\mathcal{F}:\mathcal{V}\times{\mathcal{Q}}\rightarrow\mathcal{R}^{\left\|\mathcal{A}\right\|}$ to produce a distribution over the candidate answer space $\mathcal{A}$. Generally, the entire process can be formulated as follow:
\begin{eqnarray}
\mathcal{F}(\mathcal{A}|v_i,q_i)=C(M(f_q(q_i),f_v(v_i)))
\end{eqnarray}
where the function $\mathcal{F}$ consists of an image encoder $f_v:\mathcal{V}\rightarrow\mathcal{R}^{d_v}$ to output visual features of dimension $d_v$ , a question encoder $f_q:\mathcal{Q}\rightarrow\mathcal{R}^{d_q}$ to output textual features of dimension $d_q$ , a multimodal fusion method $M:\mathcal{R}^{d_v}\times{\mathcal{R}^{d_q}} \rightarrow \mathcal{R}^{d_m}$ to output fusion features of dimension $d_m$ and a classifier $C:\mathcal{R}^{d_m}\rightarrow\mathcal{R}^{\left\|\mathcal{A}\right\|}$ to output the answer prediction $\mathcal{F}(\mathcal{A}|v_i,q_i)$ for the $i_{th}$ image and question pair.

\subsection{Classical learning strategy}

Correspondingly, the classical learning strategy of VQA models is to conduct a multi-label loss $\mathcal{L}_{ml}$, minimizing the binary cross-entropy criterion over a dataset of size $N$.
\begin{eqnarray}
\mathcal{L}_{ml}=-\frac{1}{N}\sum_i^Ny_i\log(\sigma(\mathcal{F}(\mathcal{A}|v_i,q_i))) \nonumber \\
+(1-y_i)\log(1-\sigma(\mathcal{F}(\mathcal{A}|v_i,q_i)))
\end{eqnarray}
where $\sigma(\cdot)$ denotes the sigmoid function, and $y_i$ is the label of $a_i$, denoted as $y_i\in\{0,1\}^{\left\|\mathcal{A}\right\|}$. 

\subsection{CCB learning strategy}

Our method is motivated by an intuitive cognition: for both VQA models and humans, we need to get enough content and context from multimodal information to make a decision~\cite{tang2020unbiased}. Figure 2 shows the overview of our approach, consisting of the Content branch $\mathcal{F}_{cn}$, the Context branch $\mathcal{F}_{cx}$ and a joint loss function $\mathcal{L}_{ccb}$ with language bias. We directly calculate the correlation between the question type $q_{type}$ and $\mathcal{A}$ to obtain language bias $b_i$ for the $i_{th}$ instance:
\begin{eqnarray}
b_i=P(\mathcal{A}|q_{type})
\end{eqnarray}
where $q_{type}\in\{1,2,...,64\}$ and VQA datasets divide $\mathcal{Q}$ into 64 question types. 

\begin{table*}[ht]
\begin{center}
\caption{Performance on VQA-CP v2 test and VQA v2 val} \label{tab:cap}
\begin{tabular}{c|c|cccc|cccc|c}
  \hline
   \multirow{2}{*}{Model}  &
   \multirow{2}{*}{Expl.}  & 
   \multicolumn{4}{c|}{VQA-CP v2 test$\uparrow$}& 
   \multicolumn{4}{c|}{VQA v2 val$\uparrow$}& 
   \multicolumn{1}{c}{Gap$\Delta\downarrow$}\\
   & & Overall & Yes/No & Number &Other & Overall & Yes/No & Number &Other &Overall
   
  \\
  \hline
  UpDn~\cite{anderson2018bottom} &  &39.84 & 41.96 & 12.36 & 46.26 &63.48 & 81.18 & 42.14 & 55.66 & 23.64 \\
  +SCR~\cite{wu2019self}  &   &48.47 & 70.41 & 10.42 & 47.29 &62.30 & 77.40 & 40.50 & 56.50 & 13.83\\
  +DLR~\cite{jing2020overcoming}  &   &48.87 & 70.99 & 18.72 & 45.57 &57.96 & 76.82 & 39.33 & 48.54 & 9.09\\
  +AReg~\cite{ramakrishnan2018overcoming} &   &41.17 & 65.49 & 15.48 & 35.48&62.75 &79.84 & 42.35 &55.16 & 21.58\\ 
  +Rubi~\cite{cadene2019rubi} &   &44.23 & 67.05 & 17.48 & 39.61 &61.16& - & - & - &21.93\\
  +LMH~\cite{clark2019don}  &   &52.05 & - & - & - &- & - & - & - & -\\
  +CCB  &   &\textbf{57.99} & \textbf{86.41} & \textbf{45.63} & \textbf{48.76} & 60.73& 78.37 &36.88 & 53.17 & \textbf{2.74}\\
  \hline
  +HINT & HAT & 47.70 & 70.04 & 10.68 & 46.31 &62.35 & 80.49 & 41.75 & 54.01 & 14.65\\
  +SCR & HAT & 49.17 & 71.55 & 10.72 & 47.49  &62.20 & 78.90 & 41.40 & 54.30 & 13.03\\
  +SCR & VQA-X & 49.45 & 72.36 & 10.93 & 48.02 &62.20 & 78.80 & 41.60 & 54.40 & 12.75\\
  +SSL & QICE & 57.59 & 86.53 & 29.87 & \textbf{50.03} &63.73 & - & - & - & 6.24\\
  +CSS & VQ-CSS & 58.95 & 84.37 & 49.42 & 48.21  &59.91 & 73.25 & 39.77 & 55.11 & 0.96\\
  +CCB & VQ-CSS & \textbf{59.12} & \textbf{89.12} & \textbf{51.04} & 45.62   &59.17 & 77.28 & 33.71 &52.14 & \textbf{0.05}\\
  \hline
\end{tabular}
\end{center}
\end{table*}

As common in the state-of-the-art, our base Up-Down~\cite{anderson2018bottom} model encodes the image $v_i$ as visual features $f_v(v_i)$ using the pretrained Faster RCNN network, encodes the question $q_i$ as question features $f_q(q_i)$ by Glove embeddings and GRU. Ideally, we hope a VQA model captures content from visual features and context from textual features. However, due to the current shortcoming of visual-language representation capability and the semantic complementarity of multimodal information, we construct the two branches with the same inputs.  

\textbf{Content branch.}
We build the Content branch on top of a base VQA model Up-Down~\cite{anderson2018bottom} and a typically ensemble method LMH~\cite{clark2019don}. Following the paradigm of VQA, $f_v(v_i)$ and $f_q(q_i)$ are first fed into the VQA model to capture the local key information and generate an answer distribution $\mathcal{F}(\mathcal{A}|v_i,q_i)$. Then we adopt ensemble method to reduce the effect of language bias and generate a new answer distribution as the output of our Content branch, which is denoted as:
\begin{eqnarray}
\mathcal{F}_{cn}(\mathcal{A}|v_i,q_i,b_i)=E(\mathcal{F}(\mathcal{A}|v_i,q_i),b_i)
\end{eqnarray}
Moreover, We try to further reduce the statistical prior caused by biased samples with reweighting method: 
\begin{eqnarray}
\mathcal{L}_{cn}=-\frac{1}{N}\sum_i^N(1-b_i)^r[y_i\log(\sigma(\mathcal{F}_{cn}(\mathcal{A}|v_i,q_i,b_i))) \nonumber \\
+(1-y_i)\log(1-\sigma(\mathcal{F}_{cn}(\mathcal{A}|v_i,q_i,b_i)))]
\end{eqnarray}
where $r$ is a hyperparameter and we utilize $(1-b_i)^r$ to adjust the contribution of different biased samples on the Content loss $\mathcal{L}_{cn}$.

\textbf{Context branch.}
We build an additional Context branch, in which we try to use global information to produce a evenly predicted distribution $\mathcal{F}_{cx}(\mathcal{A}|v_i,q_i)$, helping the model learn a good context prior to filter out unnecessary answer candidates. Different from the Content branch that emphasizes on obtaining the most relevant information, the Context branch obtains all information that may affect $\mathcal{F}_{cx}(\mathcal{A}|v_i,q_i)$.
\begin{eqnarray}
\mathcal{F}_{cx}(\mathcal{A}|v_i,q_i)=C_{cx}(nn_q(f_q(q_i))\odot nn_v(f_v(v_i)))
\end{eqnarray}
where $\odot$ denotes element-wise product. Concretely, $\mathcal{F}_{cx}$ uses two neural networks $nn_q:\mathcal{R}^{d_q}\rightarrow\mathcal{R}^{d_m}$, $nn_v:\mathcal{R}^{d_v}\rightarrow\mathcal{R}^{d_m}$ to project multimodal features into a common space $\mathcal{R}^{d_m}$, then their element-wise product is fed into a classifier $C_{cx}:\mathcal{R}^{d_m}\rightarrow\mathcal{R}^{\left\|\mathcal{A}\right\|}$. To learn context priors with language bias, we convert $b_i$ into a binary vector $B(b_i)$ as the label of computing $\mathcal{L}_{cx}$, denoted as follow:
\begin{eqnarray}
\mathcal{L}_{cx}=-\frac{1}{N}\sum_i^NB(b_i)\log(\sigma(\mathcal{F}_{cx}(\mathcal{A}|v_i,q_i))) \nonumber\\
+(1-B(b_i))\log(1-\sigma(\mathcal{F}_{cx}(\mathcal{A}|v_i,q_i)))
\end{eqnarray}
where $B(\cdot)$ is the function to convert $b_i$  into the label, denoted as:
\begin{eqnarray}
B(b_{ij})=\left\{\begin{aligned} 1 && b_{ij}>0 \\
0 && others \end{aligned}
\right.
\end{eqnarray}
where $b_{ij}$ is the predicted probability of $a_{ij}$, $j\in{\{1,\left\|\mathcal{A}\right\|\}}$.

\textbf{Predicting by Content and Context.} 
To obtain the final prediction $\mathcal{F}_p(\mathcal{A}|v_i,q_i,b_i)$ based on visual content and language context, we simply conduct an element-wise product between $\mathcal{F}_{cn}(\mathcal{A}|v_i,q_i,b_i)$ and $\mathcal{F}_{cx}(\mathcal{A}|v_i,q_i)$.
\begin{eqnarray}
\mathcal{F}_p(\mathcal{A}|v_i,q_i,b_i)=\mathcal{F}_{cn}(\mathcal{A}|v_i,q_i,b_i)\odot\mathcal{F}_{cx}(\mathcal{A}|v_i,q_i)
\end{eqnarray}
Follow the classical learning strategy, we conduct the Predict loss $\mathcal{L}_P$ to optimize $\mathcal{F}_p(\mathcal{A}|v_i,q_i,b_i)$ as follow:
\begin{eqnarray}
\mathcal{L}_p=-\frac{1}{N}\sum_i^Ny_i\log(\sigma(\mathcal{F}_p(\mathcal{A}|v_i,q_i,b_i))) \nonumber \\
+(1-y_i)\log(1-\sigma(\mathcal{F}_p(\mathcal{A}|v_i,q_i,b_i)))
\end{eqnarray}

At last, we obtain our final loss $\mathcal{L}_{ccb}$ by summing $\mathcal{L}_{cn}$, $\mathcal{L}_{cx}$ and $\mathcal{L}_p$ together.
\begin{eqnarray}
\mathcal{L}_{ccb}=\mathcal{L}_{cn}+\mathcal{L}_{cx}+\mathcal{L}_p
\end{eqnarray}

\section{Experiments}
\label{sec:illust}
\subsection{Setup}
We evaluate our model on the VQA v2 dataset~\cite{goyal2017making} and VQA-CP v2 dataset~\cite{agrawal2018don}, following the standard VQA evaluation metric~\cite{antol2015vqa}. For fair comparisons, we use the same method as the Up-Down~\cite{anderson2018bottom} model to extract multi-modal features and build on top of the publicly available model LMH~\cite{clark2019don} with the same settings.

\begin{figure*}[t]
\begin{minipage}[t]{1.0\linewidth}
  \centering
 \centerline{\epsfig{figure=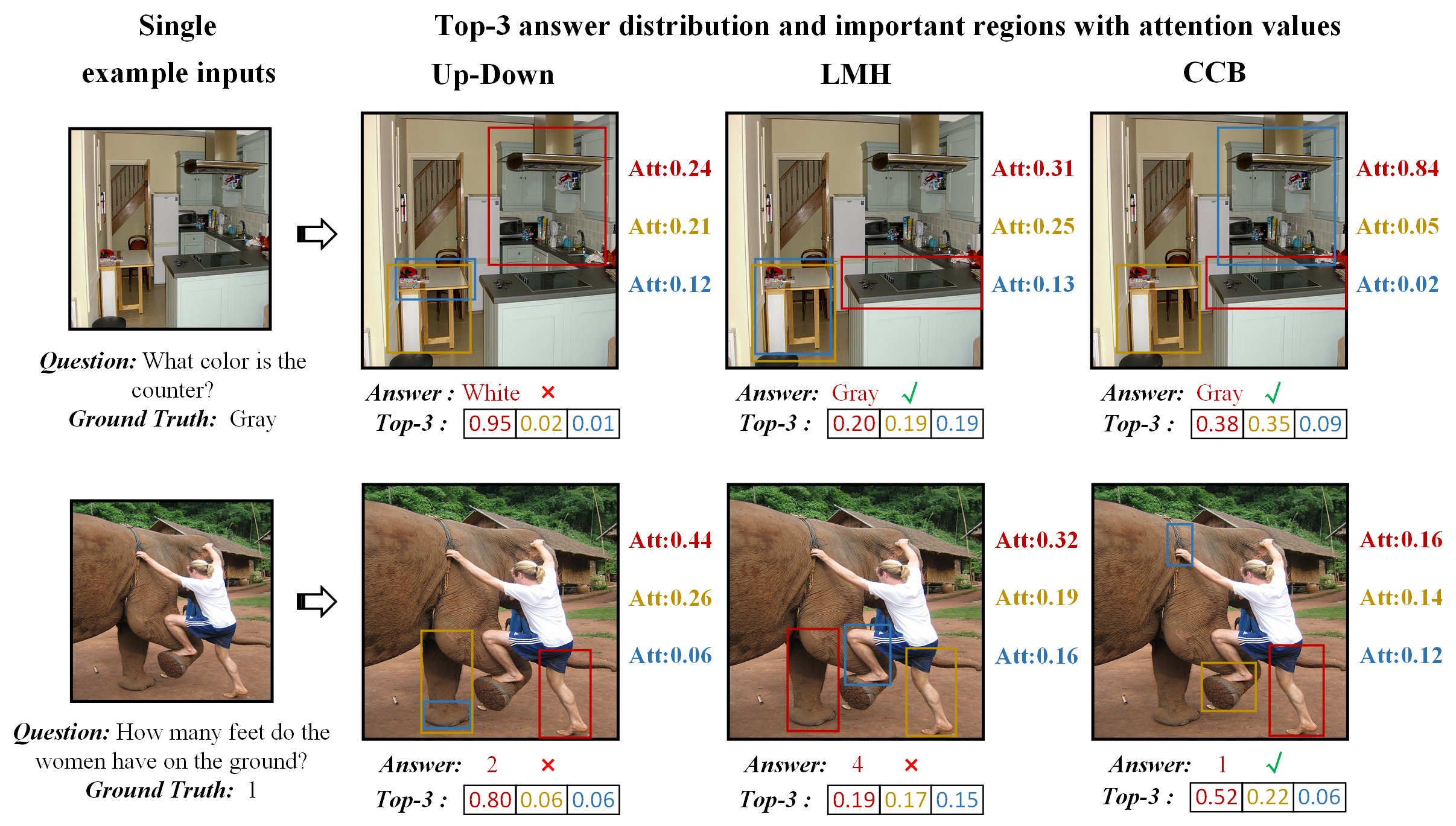,width=12cm}}
\end{minipage}
\caption{Qualitative comparison among the outputs of our baseline Up-Down model, typical ensemble method LMH and our method CCB on VQA-CP v2 test set. The bounding boxes indicate the top-3 important regions with attention values ``Att''. ``Top-3'' under ``Answer'' represents the probability of the top-3 candidate answers.}
\label{fig:res}
\end{figure*}

\subsection{Performance on VQA-CP v2 and VQA v2 datasets}

We first compare CCB with other non-extra-based methods, which are proposed to reduce language bias. Table 1 shows that our approach significantly outperforms other methods over all question categories. Particularly, without using any extra data, we improve the performance of LMH from 52.05\% to 57.99\%, which is competitive even in extra-based methods. Then, compared with some extra-based methods, CCB achieves the state-of-the-art performance with CSS training scheme, which further proves the effectiveness of our method. For some simple ``Yes/No'' and direct ``number'' questions, our method achieves higher accuracy (89.12\% and 51.04\%), while answering these two types of questions is most susceptible to statistical prior. Finally, we evaluate CCB on the biased VQA v2 dataset. As can be seen from Table 1, there is an obvious gap in the performance of most methods between these two datasets. Notably, our method effectively reduces the performance gap of the model on the two datasets and improves the robustness of the model. The reason is that our learning strategy can decrease most statistical prior on VQA-CP v2, while retaining most context prior on VQA v2 so as to avoid a huge drop in performance. 

\subsection{Ablation Studies}

Ablation studies on the VQA-CP v2 dataset are conducted to investigate factors that affect the performance of our approach. The results are listed in Table 2. Firstly we investigate the effectiveness of building Content and Context branches. To this end, we set ``$r=0$, w/o'', in which the Content branch uses a  multi-label loss $\mathcal{L}_{ml}$ and the Context branch calculates $\mathcal{L}_{cx}$ by setting an all-one constant vector as the label. The ``$r=0$, w/o'' performs better than the LMH model, which indicates that the two branches can jointly affect the generation of answers avoiding only depending on one-sided information. Then we set $r=1$ and add the context label to investigate the influence of using bias for reweighting in $\mathcal{L}_{cn}$ and labeling in $\mathcal{L}_{cx}$ respectively. The overall accuracy of 55.70\% and 56.76\% demonstrate that our loss functions are effective to reduce the effect of biased samples as well as retain their beneficial influence on learning prior knowledge. Finally, we investigate the influence of different hyper-parameter  $r$, which are applied to adjust the importance of biased samples. The results show that the best performance can be achieved when $r=1$. For further analysis, a large $r$ may affect the model's learning ability to biased samples, while a small $r$ may make the model get no complete rid of the excessive dependence on language bias.

\begin{table}[t]
\begin{center}
\caption{Ablation Studies on VQA-CP v2 test} \label{tab:cap}
\begin{tabular}{cccc}
  \hline
  Up-Down & $(1-bias)^r$ & context label & Accuracy
  \\
  \hline
  +None & -& -&39.68\\
  +LMH~\cite{clark2019don} & -& -& 52.05 \\
  +CCB  &$r=0$ & w/o&55.06 \\
  +CCB  & $r=1$& w/o & 55.70 \\
  +CCB  & $r=0$& w& 56.76 \\
  +CCB  &$r=1$ & w& \textbf{57.99} \\
  +CCB  & $r=0.5$& w& 57.39 \\
  +CCB  &$r=2$& w&57.56 \\
  \hline
\end{tabular}
\end{center}
\end{table}

\subsection{Qualitative analysis}

We make qualitative evaluation of the effectiveness of our method by visualizing the top-3 important regions of three models and outputting the corresponding attention weights. In addition, we apply the softmax function to calculate the probability of top-3 candidate answers predicted by the models, which reflects the models' confidence in the answer. As shown in Figure 3, for the question like ``What color is the counter ?'', although both LMH and our model give the correct answer, our method makes the model pay more attention to ``the counter'' and ignore other regions that are not related to the question. This allows our model to give a more accurate color of ``the counter''. Even for the question that requires more understanding of visual content like ``How many feet do the women have on the ground ?'', our model can still give the correct answer. It can be seen that, the model correctly applies more weights to the corresponding regions in the image, including ``the feet of the woman'' on the ground and ``other parts of the body'' that are in contact with ``the elephant''. In both cases, our method can avoid the model from being directly affected by the statistical prior and giving common answers ({\em e.g.}, ``white'', ``2''). Moreover, our method can guide the model to understand the whole question and apply the correct attention weights , instead of just focusing on the n-gram ``How many feet ...''.

\section{Conclusion}

We propose CCB learning strategy to cope with language bias for Visual Question Answering (VQA) task. Based on the Content and the Context branches, CCB guides the model to answer questions with the combination of decisive content information and necessary context information. Furthermore, we construct an additional loss function to jointly optimize the two branches and the final prediction by disentangling the influence of language bias on the model. Experimental results show that our approach makes a balance between reducing statistical prior and preserving context prior, and the state-of-the-art performance is achieved on VQA-CP v2 dataset.

\bibliographystyle{IEEEbib}
\bibliography{CCB}

\begin{thebibliography}{10}

\bibitem{antol2015vqa}
S. Antol, A. Agrawal, J. Lu, M. Mitchell, D. Batra,
  C~Lawrence~Zitnick, and D. Parikh,
\newblock ``Vqa: Visual question answering,''
\newblock in {\em ICCV}, 2015, pp. 2425--2433.

\bibitem{goyal2017making}
Y. Goyal, T. Khot, D. Summers-Stay, D. Batra, and D. Parikh,
\newblock ``Making the v in vqa matter: Elevating the role of image
  understanding in visual question answering,''
\newblock in {\em CVPR}, 2017, pp. 6904--6913.

\bibitem{agrawal2018don}
A. Agrawal, D. Batra, D. Parikh, and A. Kembhavi,
\newblock ``Don't just assume; look and answer: Overcoming priors for visual
  question answering,''
\newblock in {\em CVPR}, 2018, pp. 4971--4980.

\bibitem{niu2020counterfactual}
Y. Niu, K. Tang, H. Zhang, Z. Lu, X. Hua, and Ji-Rong
  Wen,
\newblock ``Counterfactual vqa: A cause-effect look at language bias,''
\newblock {\em arXiv preprint arXiv:2006.04315}, 2020.

\bibitem{ramakrishnan2018overcoming}
S. Ramakrishnan, A. Agrawal, and S. Lee,
\newblock ``Overcoming language priors in visual question answering with
  adversarial regularization,''
\newblock in {\em NIPS}, 2018, pp.
  1541--1551.

\bibitem{yang2016stacked}
Z. Yang, X. He, J. Gao, L. Deng, and A. Smola,
\newblock ``Stacked attention networks for image question answering,''
\newblock in {\em CVPR}, 2016, pp. 21--29.

\bibitem{anderson2018bottom}
P. Anderson, X. He, C. Buehler, D. Teney, M. Johnson, S.
  Gould, and L. Zhang,
\newblock ``Bottom-up and top-down attention for image captioning and visual
  question answering,''
\newblock in {\em CVPR}, 2018, pp. 6077--6086.

\bibitem{selvaraju2019taking}
R. Selvaraju, S. Lee, Y. Shen, H. Jin, S. Ghosh,
  L. Heck, D. Batra, and D. Parikh,
\newblock ``Taking a hint: Leveraging explanations to make vision and language
  models more grounded,''
\newblock in {\em ICCV}, 2019, pp. 2591--2600.

\bibitem{wu2019self}
J. Wu and R. Mooney,
\newblock ``Self-critical reasoning for robust visual question answering,''
\newblock in {\em NIPS}, 2019, pp.
  8604--8614.

\bibitem{qiao2020rankvqa}
Y. Qiao, Z. Yu, and J. Liu,
\newblock ``Rankvqa: Answer re-ranking for visual question answering,''
\newblock in {\em ICME}. IEEE, 2020, pp. 1--6.

\bibitem{zhu2020overcoming}
X. Zhu, Z. Mao, C. Liu, P. Zhang, B. Wang, and Y. Zhang,
\newblock ``Overcoming language priors with self-supervised learning for visual
  question answering,''
\newblock in {\em IJCAI}, 2020, pp. 1083--1089

\bibitem{cadene2019rubi}
R. Cadene, C. Dancette, M. Cord, D. Parikh, {\em et~al.},
\newblock ``Rubi: Reducing unimodal biases for visual question answering,''
\newblock in {\em NIPS}, 2019, pp.
  841--852.

\bibitem{clark2019don}
C. Clark, M. Yatskar, and L. Zettlemoyer,
\newblock ``Don't take the easy way out: Ensemble based methods for avoiding
  known dataset biases,''
\newblock {\em arXiv preprint arXiv:1909.03683}, 2019.

\bibitem{chen2020counterfactual}
L. Chen, X. Yan, J. Xiao, H. Zhang, S. Pu, and Y. Zhuang,
\newblock ``Counterfactual samples synthesizing for robust visual question
  answering,''
\newblock in {\em CVPR}, 2020, pp. 10800--10809.

\bibitem{das2017human}
A. Das, H. Agrawal, L. Zitnick, D. Parikh, and D. Batra,
\newblock ``Human attention in visual question answering: Do humans and deep
  networks look at the same regions?,''
\newblock {\em Computer Vision and Image Understanding}, vol. 163, pp. 90--100,
  2017.

\bibitem{huk2018multimodal}
D. H.Park, L. A. Hendricks, Z. Akata, A. Rohrbach, B. Schiele,
  T. Darrell, and M. Rohrbach,
\newblock ``Multimodal explanations: Justifying decisions and pointing to the
  evidence,''
\newblock in {\em CVPR}, 2018, pp. 8779--8788.

\bibitem{tang2020unbiased}
K. Tang, Y. Niu, J. Huang, J. Shi, and H. Zhang,
\newblock ``Unbiased scene graph generation from biased training,''
\newblock in {\em CVPR}, 2020, pp. 3716--3725.

\bibitem{grand2019adversarial}
G. Grand and Y. Belinkov,
\newblock ``Adversarial regularization for visual question answering:
  Strengths, shortcomings, and side effects,''
\newblock {\em arXiv preprint arXiv:1906.08430}, 2019.

\bibitem{jing2020overcoming}
C. Jing, Y. Wu, X. Zhang, Y. Jia, and Q. Wu,
\newblock ``Overcoming language priors in vqa via decomposed linguistic
  representations.,''
\newblock in {\em AAAI}, 2020, pp. 11181--11188.

\end{thebibliography}

\end{document}